\PassOptionsToPackage{prologue,dvipsnames}{xcolor}

\documentclass[sigconf,authorversion=true,nonacm=true]{acmart}

\usepackage{ifthen}
\usepackage[american]{babel}
\usepackage{graphicx}
\usepackage{float}
\usepackage[position=b]{subcaption}
\usepackage{booktabs}
\usepackage{siunitx}
\usepackage{listings}
\usepackage{nicefrac}
\usepackage[inline]{enumitem}
\usepackage{siunitx}
\usepackage{stmaryrd}
\usepackage{threeparttable}
\usepackage{multirow}
\usepackage{lipsum}
\usepackage{soul}
\usepackage[capitalise,noabbrev]{cleveref}
\usepackage{doi}

\creflabelformat{equation}{#2\textup{#1}#3}
\crefname{lstfloat}{Listing}{Listings}
\crefname{sublstfloat}{Listing}{Listings}

\definecolor{Palette1}{rgb}{1.0, 0.0, 0.0}
\definecolor{Palette2}{rgb}{0.0, 0.22, 0.66}
\definecolor{Palette3}{rgb}{0.85, 0.65, 0.13}

\usepackage{tikz}
\usepackage{ifthen}

\usetikzlibrary{decorations.pathmorphing,decorations.pathreplacing,calligraphy}
\usetikzlibrary{shapes,shapes.misc,arrows,positioning,calc,arrows.meta}

\tikzset{snake it/.style={decorate, decoration={snake, segment length=3mm, amplitude=0.5mm}}}

\tikzset{drawing layout large/.style={%
    y=-2.5mm,%
    x=2.5mm,%
    every node/.style={%
        circle, fill=black, scale=0.25%
    },%
    every path/.style={%
        gray, thick, shorten >=0.25mm, shorten <=0.25mm, -{Stealth[width=4pt,length=3pt]}
    }%
}}

\tikzset{drawing layout/.style={%
    y=-5mm,
    x=5mm,
    every node/.style={%
        circle, fill=black, scale=0.25
    },
    every path/.style={%
        gray, thick, shorten >=0.7mm, shorten <=0.7mm, -{Stealth[width=4pt,length=3pt]}
    }
}}

\tikzset{drawing layout small/.style={%
    y=-5mm,
    x=2.5mm,
    every node/.style={%
        circle, fill=black, scale=0.25
    },
    every path/.style={%
        gray, thick, shorten >=0.2mm, shorten <=0.2mm, -{Stealth[width=4pt,length=3pt]}
    }
}}

\lstdefinestyle{mystyle}{%
    xrightmargin=5pt,%
    xleftmargin=10pt,%
    numberstyle=\tiny\color{gray},%
    basicstyle=\textup\ttfamily\scriptsize,%
    breakatwhitespace=false,%
    breaklines=true,%
    captionpos=b,%
    keepspaces=true,%
    numbers=left,%
    numbersep=5pt,%
    showspaces=false,%
    showstringspaces=false,%
    showtabs=false,%
    tabsize=2%
}

\makeatletter
\newcommand{\blind}[1]{%
  \if@ACM@anonymous%
    \hl{[redacted for review]}%
  \else%
    #1%
  \fi%
}
\makeatother

\newfloat{lstfloat}{t}{lop}
\floatname{lstfloat}{Listing}

\DeclareCaptionSubType{lstfloat}

\newcolumntype{R}[1]{>{\raggedleft\let\newline\\\arraybackslash\hspace{0pt}}m{#1}}

\newcommand{\curvesmall}[1]{\includegraphics{tmp_figures/layout_small_#1}}
\newcommand{\curve}[1]{\includegraphics{tmp_figures/layout_large_#1}}

\usepackage{tikz}
\usepackage{ifthen}

\usetikzlibrary{decorations.pathmorphing,decorations.pathreplacing,calligraphy}
\usetikzlibrary{shapes,shapes.misc,arrows,positioning,calc,arrows.meta}

\tikzset{snake it/.style={decorate, decoration={snake, segment length=3mm, amplitude=0.5mm}}}

\tikzset{drawing layout large/.style={%
    y=-2.5mm,%
    x=2.5mm,%
    every node/.style={%
        circle, fill=black, scale=0.25%
    },%
    every path/.style={%
        gray, thick, shorten >=0.25mm, shorten <=0.25mm, -{Stealth[width=4pt,length=3pt]}
    }%
}}

\tikzset{drawing layout/.style={%
    y=-5mm,
    x=5mm,
    every node/.style={%
        circle, fill=black, scale=0.25
    },
    every path/.style={%
        gray, thick, shorten >=0.7mm, shorten <=0.7mm, -{Stealth[width=4pt,length=3pt]}
    }
}}

\tikzset{drawing layout small/.style={%
    y=-5mm,
    x=2.5mm,
    every node/.style={%
        circle, fill=black, scale=0.25
    },
    every path/.style={%
        gray, thick, shorten >=0.2mm, shorten <=0.2mm, -{Stealth[width=4pt,length=3pt]}
    }
}}

\begin{document}
\title{Using Evolutionary Algorithms to Find Cache-Friendly Generalized Morton Layouts for Arrays}

\author{Stephen Nicholas Swatman}
\orcid{0000-0002-3747-3229}
\affiliation{%
  \institution{University of Amsterdam}
  \city{Amsterdam}
  \country{The Netherlands}
}
\additionalaffiliation{%
    \institution{CERN}
    \city{Geneva}
    \country{Switzerland}
}
\email{s.n.swatman@uva.nl}

\author{Ana-Lucia Varbanescu}
\orcid{0000-0002-4932-1900}
\affiliation{%
  \institution{University of Twente}
  \city{Enschede}
  \country{The Netherlands}
}
\email{a.l.varbanescu@utwente.nl}

\author{Andy D. Pimentel}
\orcid{0000-0002-2043-4469}
\affiliation{%
  \institution{University of Amsterdam}
  \city{Amsterdam}
  \country{The Netherlands}
}
\email{a.d.pimentel@uva.nl}

\author{Andreas Salzburger}
\orcid{0000-0001-6004-3510}
\affiliation{%
    \institution{CERN}
    \city{Geneva}
    \country{Switzerland}
}
\email{andreas.salzburger@cern.ch}

\author{Attila Krasznahorkay}
\orcid{0000-0002-6468-1381}
\affiliation{%
    \institution{CERN}
    \city{Geneva}
    \country{Switzerland}
}
\email{attila.krasznahorkay@cern.ch}



\begin{abstract}
The layout of multi-dimensional data can have a significant impact on the efficacy of hardware caches and, by extension, the performance of applications. Common multi-dimensional layouts include the canonical row-major and column-major layouts as well as the Morton curve layout. In this paper, we describe how the Morton layout can be generalized to a very large family of multi-dimensional data layouts with widely varying performance characteristics. We posit that this design space can be efficiently explored using a combinatorial evolutionary methodology based on genetic algorithms. To this end, we propose a chromosomal representation for such layouts as well as a methodology for estimating the fitness of array layouts using cache simulation. We show that our fitness function correlates to kernel running time in real hardware, and that our evolutionary strategy allows us to find candidates with favorable simulated cache properties in four out of the eight real-world applications under consideration in a small number of generations. Finally, we demonstrate that the array layouts found using our evolutionary method perform well not only in simulated environments but that they can effect significant performance gains---up to a factor ten in extreme cases---in real hardware.
\end{abstract}

\maketitle

\renewcommand{\shortauthors}{Stephen Nicholas Swatman, Ana-Lucia Varbanescu, et al.}
\section{Introduction}

Structured multi-dimensional data are ubiquitous in high-per\-for\-mance computing~\cite{10.1145/1562764.1562783}: three-dimensional fluid simulations, dense linear algebra operations, and stencil kernels are just a few examples of applications which rely fundamentally on multi-dimensional arrays. In spite of the importance of such applications, however, most modern computer systems have one-dimensional memories: from the perspective of the programmer, memory is nothing more than a very large one-dimensional array of bytes. This discrepancy between application requirements and hardware design requires programmers to carefully consider \emph{array layouts}: injective functions which translate multi-dimensional indices into one-dimensional memory addresses.

Although array layouts do not impact the functional properties of programs, choosing a suitable layout can significantly impact application performance in modern processors with complex cache hierarchies~\cite{1214317}. Exploiting these caches is of critical importance to achieving high performance in all but purely compute-bound applications, but doing so requires locality of access---both temporal and spatial---in memory. Kernels often exhibit locality in multiple dimensions, and a well-chosen array layout maximizes the degree to which this application-level locality is translated to the address-level locality that caches are designed to exploit; as a result, that layout increases the efficacy of hardware caching and---by extension---the performance of an application.

Data in two-dimensions is commonly laid out in \emph{row-major} order (shown in \cref{fig:layouts8x8:000111} for an $8 \times 8$ array) or \emph{column-major} order (\cref{fig:layouts8x8:111000}) which provide good locality of access in a single dimension, but poor locality in all others. Thankfully, the design space for data orderings---in two dimensions or more---extends far beyond these canonical layouts: the \emph{Morton} layout (\cref{fig:layouts8x8:010101}), for example, is a layout based on a space-filling curve which provides balanced locality between multiple dimensions~\cite{morton1966computer,10.1002/cpe.1018}. Our work explores a family of data layouts which generalize the Morton order, and allow us to carefully tune the cache behavior in multiple dimensions to match a given application.

The design space of the aforementioned family of data layouts is dauntingly large; indeed, the number of possible layout for arrays at scales applicable to real-world problems is so large that it renders exhaustive search infeasible. In order to find suitable array layouts in tractable amounts of time, we propose to employ genetic algorithms---heuristics known to be able to efficiently find high-quality solutions in large search spaces~\cite{holland1992genetic}. To this end, we design a chromosomal representation of Morton-like array layouts, as well as a fitness function that uses cache simulation to estimate the performance of individual array layouts. Finally, we evaluate our evolutionary strategy and the array layouts it discovers.

In short, our paper makes the following contributions:

\begin{itemize}
    \item We characterize the design space given by a generalization of the Morton array layout, and we show that that the size of this design space renders exhaustive search infeasible (\cref{sec:bijections});
    \item We propose an evolutionary methodology based on genetic algorithms for exploring the aforementioned design space based on the simulated cache-friendliness of layouts (\cref{sec:exploration});
    \item We design and execute a series of experiments to assess the accuracy of our fitness function, the efficacy of our evolutionary process, and the performance of the discovered array layouts, showing that our method can improve performance up to a factor ten (\cref{sec:experiments}).
\end{itemize}

\section{Background and Related Work}

\label{sec:background}

In this section, we provide a brief overview of the basic concepts and notations which are essential to the remainder of this paper, and highlight relevant related work. 

\subsection{Indexing Functions and Canonical Layouts}

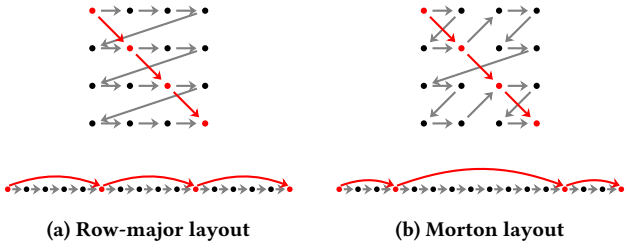
\begin{figure}
    \centering
    \begin{subfigure}{0.48\columnwidth}
        \centering
        \curvesmall{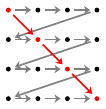}\\
        \begin{tikzpicture}[drawing layout small]
            \foreach \x in {0,...,15} {
                \ifnum\ifnum\x=0 1\else\ifnum\x=5 1\else\ifnum\x=10 1\else\ifnum\x=15 1\else0\fi\fi\fi\fi
                =1 %
                \node (p\x) at (\x,0) [fill=Palette1] {};
                \else
                \node (p\x) at (\x,0) {};
                \fi
            }

            \foreach \x in {0,...,14} {
                \pgfmathtruncatemacro{\nx}{\x+1}
                \xdef\nx{\nx}
                \draw (p\x) -> (p\nx);
            }

            \path [draw=none] (current bounding box.south west) +(-1mm,-1mm) (current bounding box.north east) +(1mm,6mm);
            \draw[Palette1] (p0.north) to [out=20, in=160] (p5.north);
            \draw[Palette1] (p5.north) to [out=20, in=160] (p10.north);
            \draw[Palette1] (p10.north) to [out=20, in=160] (p15.north);
        \end{tikzpicture}
        \caption{Row-major layout}
        \label{fig:locality_example:row_major}
    \end{subfigure}\hfill
    \begin{subfigure}{0.48\columnwidth}
        \centering
        \curvesmall{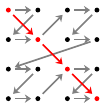}\\
        \begin{tikzpicture}[drawing layout small]
            \foreach \x in {0,...,15} {
                \ifnum\ifnum\x=0 1\else\ifnum\x=3 1\else\ifnum\x=12 1\else\ifnum\x=15 1\else0\fi\fi\fi\fi
                =1 %
                \node (p\x) at (\x,0) [fill=Palette1] {};
                \else
                \node (p\x) at (\x,0) {};
                \fi
            }

            \foreach \x in {0,...,14} {
                \pgfmathtruncatemacro{\nx}{\x+1}
                \xdef\nx{\nx}
                \draw (p\x) -> (p\nx);
            }

            \path [draw=none] (current bounding box.south west) +(-1mm,-1mm) (current bounding box.north east) +(1mm,6mm);
            \draw[Palette1] (p0.north) to [out=20, in=160] (p3.north);
            \draw[Palette1] (p3.north) to [out=20, in=160] (p12.north);
            \draw[Palette1] (p12.north) to [out=20, in=160] (p15.north);
        \end{tikzpicture}
        \caption{Morton layout}
        \label{fig:locality_example:morton}
    \end{subfigure}
    \caption{Two-dimensional arrays laid out in memory along the gray arrows. An application accesses the array diagonally along the red arrows. Application locality is shown above, memory locality is shown below.}
    \label{fig:locality_example}
\end{figure}

Dense $n$-dimensional arrays can be imagined as structured grids in which each element is assigned to exactly one point in $\mathbb{N}^n$. In most modern processors, multi-dimensional arrays are a software-level abstraction over the one-dimensional memory of the machine; in order to actually access multi-dimensional data, we need to define a function that converts indices in $n$ dimensions to memory addresses\footnote{In reality, address calculations must also consider array offsets (the address of the first element) and scales (the size of each element). We skip over these complications as they are handled transparently by address generation units in modern hardware, and they affect all array layouts in the same manner.}. We refer to the class of such functions as \emph{indexing functions}, and they are isomorphic to \emph{array layouts}. In short, an $n$-dimensional indexing functions is an injective (often bijective) function of the following type, where $N_i$ represents the size of the array in the $i$th dimension, $\bigtimes$ is the generalised Cartesian product, and $\llbracket a, b\rrbracket$ is the integer interval from $a$ to $b$:

\begin{equation}
f : \bigtimes_{i=0}^{n-1} \llbracket 0,N_i - 1 \rrbracket \to \left\llbracket0,\left(\prod_{i=0}^{n-1} N_i\right) - 1 \right\rrbracket
\end{equation}

In a multi-dimensional grid, we denote the elements along a given axis---that is to say, the sequence of elements for which all indices except one are fixed---as \emph{fibers}~\cite{doi:10.1137/07070111X}. In a two-dimensional case, fibers along the $x$-axis are known as \emph{rows}, and fibers along the $y$-axis as columns. In order to facilitate the description of arrays in three or more dimensions, we use the term \emph{mode}-$m$ fibers to describe fibers along the $m$th dimension, such that mode-0 fibers are synonymous with rows, mode-1 fibers refer to columns, and so forth.

The most common group of multi-dimensional indexing functions are the \emph{canonical} layouts, sometimes known as the \emph{lexicographic} layouts or, in the two-dimensional case, the \emph{row-} and \emph{column-major} layouts. In a canonical layout, one-dimensional array indices are calculated according to \cref{eq:lexicographic}, in which $x_0, \ldots, x_{n-1}$ are components of the $n$-dimensional index, and $N_0, \ldots, N_{n-1}$ represent the size of the array in each dimension:

\begin{equation}
\label{eq:lexicographic}
f(x_0, \ldots, x_{n-1}; N_0, \ldots, N_{n-1}) = \sum_{i=0}^{n-1}\left(\prod_{j=0}^{i-1}N_j\right)x_i
\end{equation}

An important corollary of \cref{eq:lexicographic} is that the mode-0 fibers are contiguous in memory i.e., \cref{eq:lexicographicconsec} holds:

\begin{equation}
\label{eq:lexicographicconsec}
f(x_0 + 1, x_1, \ldots, x_{n-1}) = f(x_0, x_1, \ldots, x_{n-1}) + 1
\end{equation}

It is worth noting that the calculation of addresses in column-major layout---in which the mode-1 fibers are contiguous---is also given by \cref{eq:lexicographic}, with the order of the indices and sizes swapped. The canonical array layouts achieve perfect spatial locality in one dimension: if a kernel accesses memory along mode-$m$ fibers, then a canonical layout where the $m$th dimension is major will provide the optimal translation between locality in the multi-dimensional space to locality in memory. Many real world applications, however, exhibit locality in multiple dimensions; a kernel might, for example, iterate diagonally over an array; an example of this---and the resulting locality in memory---is given in \cref{fig:locality_example:row_major}.

The performance of canonical storage layouts has been studied extensively.
\citeauthor{1214317} discuss methods for compensating for the weaknesses of canonical layouts using tiling and recursive layouts~\cite{1214317}.
Similarly, \citeauthor{kowarschik2003overview} propose a variety of strategies that mitigate cache misses in canconical storage layouts for numerical applications~\cite{kowarschik2003overview}.
\citeauthor{1560002} propose a metric for the locality of array layouts~\cite{1560002}.
\citeauthor{5473222} analyze the performance of access patterns in multi-dimensional data in graphics processing units (GPUs)~\cite{5473222}. \citeauthor{10.1145/2063384.2063401} propose a method for automatically optimizing storage layouts~\cite{10.1145/2063384.2063401}.

\subsection{Morton Layouts}

The Morton order is a notable example of a non-canonical array layout that provides balanced locality in multiple dimensions. It is conceptually simple to understand, efficient to implement in commodity hardware (as we will show in \cref{sec:bijections:accel}), and it has been shown to positively affect the efficacy of hardware caches: \citeauthor{doi:10.1177/1094342019846282} show the efficacy of the Morton layout in molecular dynamics applications~\cite{doi:10.1177/1094342019846282},  \citeauthor{9378385} describe its benefits in matrix decomposition~\cite{9378385}, and \citeauthor{10.1002/cpe.1018} provide an in-depth performance analysis of this array layout in a range of kernels~\cite{10.1002/cpe.1018}. \citeauthor{10.1145/305619.305645} show the applicability of Morton layouts---as well as other non-canonical layouts---in matrix multiplication~\cite{10.1145/305619.305645}, and this work is expanded upon in~\cite{10.1145/305138.305231}. Applications of the Morton order in more than two dimensions have been studied by \citeauthor{PAWLOWSKI201934}~\cite{PAWLOWSKI201934}. \citeauthor{Mellor-Crummey2001} show the applicability of array layouts based on space-filling curves---like the Morton layout---for irregular applications~\cite{Mellor-Crummey2001}. The practical applicability of the Morton layout is further evidenced by the \textsc{Opie} compiler, which employs Morton array layouts natively~\cite{10.1145/1054943.1054962}.

The performance benefits of the Morton layout stem from its spatial structure: an example---which justifies why this layout is sometimes known as the \emph{Z-order layout}---is given in \cref{fig:locality_example:morton}; note the difference in locality in the address space compared to the canonical layout (\cref{fig:locality_example:row_major}). The Morton order layout has also been applied to data movement in parallel systems by \citeauthor{walker2023impact}~\cite{walker2023impact}, and \citeauthor{6687350} have applied the layout to workload distribution in parallel processes~\cite{6687350}. \citeauthor{bader2012space} explores a variety of applications of space-filling curves in scientific programs~\cite{bader2012space}. \citeauthor{10.14778/3415478.3415560} explore the application of Morton curves for the storage of databases, reducing the total amount of data read from persistent storage~\cite{10.14778/3415478.3415560}; although the aforementioned paper considers a much higher level of abstraction than the methods in this paper---which operate at the level of hardware caches---we believe that the methods presented in this paper may generalize to a broader range of applications, including databases.

In the Morton order, multi-dimensional indices can be converted to one-dimensional addresses in a variety of ways. The Moser--de Bruijn sequence~\cite{mosedebruijn} is commonly used as it allows efficient conversions in two dimensions, but this method requires us to store the Moser--de Bruijn sequence in memory, and accessing this sequence causes additional overhead. Therefore, we prefer a different method based on the interleaving of the (unsigned) binary representation of multi-dimensional indices. As an example, the two-dimensional index $(3, 5)$ can be bijectively mapped into one-dimensional memory by finding the binary expansions of the indices i.e., $({\color{Palette1}011}_2, {\color{Palette2}101}_2)$, and interleaving the bits yielding ${\color{Palette2}1}{\color{Palette1}0}{\color{Palette2}0}{\color{Palette1}1}{\color{Palette2}1}{\color{Palette1}1}_2 = 39_{10}$.  This is equivalent to first dilating and shifting the binary expansions of the numbers, and then taking their bitwise disjunction (OR): the first index is dilated yielding ${\color{gray}0}{\color{Palette1}0}{\color{gray}0}{\color{Palette1}1}{\color{gray}0}{\color{Palette1}1}_2$ while the second index is dilated and shifted left yielding ${\color{Palette2}1}{\color{gray}0}{\color{Palette2}0}{\color{gray}0}{\color{Palette2}1}{\color{gray}0}_2$. Taking the bitwise disjunction of these numbers yields the same address as using the interleaving strategy. The computation of Morton indices through bit manipulation can be extended to an arbitrary number of dimensions; the three-dimensional index $(3, 5, 4)$ expands to $({\color{Palette1}011}_2, {\color{Palette2}101}_2, {\color{Palette3}100}_2)$, and the resulting memory address is ${\color{Palette3}1}{\color{Palette2}1}{\color{Palette1}0}{\color{Palette3}0}{\color{Palette2}0}{\color{Palette1}1}{\color{Palette3}0}{\color{Palette2}1}{\color{Palette1}1}_2 = 395_{10}$. Note that the relative significance of bits in each of the input indices is preserved in the output address. \citeauthor{10.1145/1274971.1274989} present the idea that the Morton layout can be generalized by allowing arbitrary bit-interleaving orders~\cite{10.1145/1274971.1274989,doi:10.1080/17445760902758560}, which is foundational to our work. This idea is further expanded on by \citeauthor{doi:10.1177/1094342017725568}~\cite{doi:10.1177/1094342017725568}.

\subsection{Genetic Algorithms}

\label{sec:backgroun:genetic}

Genetic algorithms are a class of heuristics introduced by \citeauthor{holland1992adaptation} which are designed to solve optimization and search problems by emulating the process of evolution as it happens in the natural world~\cite{holland1992adaptation}. In genetic algorithms, \emph{generations} of \emph{individuals} i.e., sets of candidate solutions, iteratively explore a design space through genetic operators. In particular, \emph{crossover} operators model the combination of the genetic material of two or more individuals~\cite{10.1145/3009966}, and \emph{mutation} operators model random changes to the gene pool~\cite{294849}. In genetic algorithms, individuals are removed from the population based on their \emph{fitness} i.e., the quality of the solution they represent to the problem posed~\cite{Sivanandam2008}. Genetic algorithms have seen successful application in an extremely broad range of fields, ranging from drug discovery~\cite{TERFLOTH2001102} to music composition~\cite{doi:10.1080/0749446032000150870}. Genetic algorithms have also proven useful for design space exploration in computer systems; \citeauthor{7738470} shows that they can be applied in the design of embedded systems~\cite{7738470}. \citeauthor{10.1007/978-3-030-55789-8_61} show that a broader class of evolutionary approaches can be used in the design of neural networks~\cite{10.1007/978-3-030-55789-8_61}. The optimization problem we consider in this paper is combinatorial in nature, and the application of genetic algorithms to such problems has also been extensively studied and proven across a variety of domains~\cite{doi:10.1287/ijoc.6.2.161,10.1007/3-540-55027-5_23,hegerty2009comparative,Goncalves2011}

\section{Generalized Morton Layouts}

\label{sec:bijections}

\begin{figure}
\begin{subfigure}{0.25\columnwidth}\centering\curve{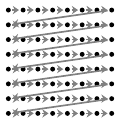}\vspace{-1.5mm}\caption{[0,0,0,1,1,1]}\vspace{1.5mm}\label{fig:layouts8x8:000111}\end{subfigure}\hfill
\begin{subfigure}{0.25\columnwidth}\centering\curve{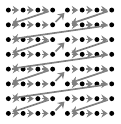}\vspace{-1.5mm}\caption{[0,0,1,0,1,1]}\vspace{1.5mm}\end{subfigure}\hfill
\begin{subfigure}{0.25\columnwidth}\centering\curve{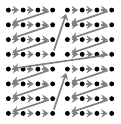}\vspace{-1.5mm}\caption{[0,0,1,1,0,1]}\vspace{1.5mm}\end{subfigure}\hfill
\begin{subfigure}{0.25\columnwidth}\centering\curve{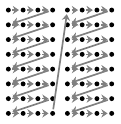}\vspace{-1.5mm}\caption{[0,0,1,1,1,0]}\vspace{1.5mm}\end{subfigure}\\
\begin{subfigure}{0.25\columnwidth}\centering\curve{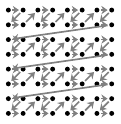}\vspace{-1.5mm}\caption{[0,1,0,0,1,1]}\vspace{1.5mm}\end{subfigure}\hfill
\begin{subfigure}{0.25\columnwidth}\centering\curve{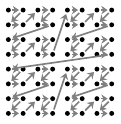}\vspace{-1.5mm}\caption{[0,1,0,1,0,1]}\vspace{1.5mm}\label{fig:layouts8x8:010101}\end{subfigure}\hfill
\begin{subfigure}{0.25\columnwidth}\centering\curve{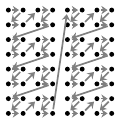}\vspace{-1.5mm}\caption{[0,1,0,1,1,0]}\vspace{1.5mm}\end{subfigure}\hfill
\begin{subfigure}{0.25\columnwidth}\centering\curve{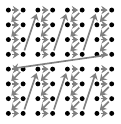}\vspace{-1.5mm}\caption{[0,1,1,0,0,1]}\vspace{1.5mm}\end{subfigure}\\
\begin{subfigure}{0.25\columnwidth}\centering\curve{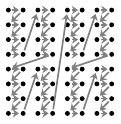}\vspace{-1.5mm}\caption{[0,1,1,0,1,0]}\vspace{1.5mm}\end{subfigure}\hfill
\begin{subfigure}{0.25\columnwidth}\centering\curve{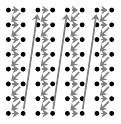}\vspace{-1.5mm}\caption{[0,1,1,1,0,0]}\vspace{1.5mm}\end{subfigure}\hfill
\begin{subfigure}{0.25\columnwidth}\centering\curve{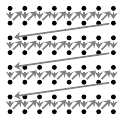}\vspace{-1.5mm}\caption{[1,0,0,0,1,1]}\vspace{1.5mm}\end{subfigure}\hfill
\begin{subfigure}{0.25\columnwidth}\centering\curve{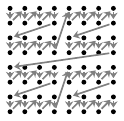}\vspace{-1.5mm}\caption{[1,0,0,1,0,1]}\vspace{1.5mm}\end{subfigure}\\
\begin{subfigure}{0.25\columnwidth}\centering\curve{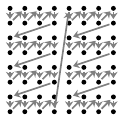}\vspace{-1.5mm}\caption{[1,0,0,1,1,0]}\vspace{1.5mm}\label{fig:layouts8x8:100110}\end{subfigure}\hfill
\begin{subfigure}{0.25\columnwidth}\centering\curve{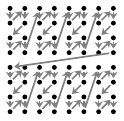}\vspace{-1.5mm}\caption{[1,0,1,0,0,1]}\vspace{1.5mm}\end{subfigure}\hfill
\begin{subfigure}{0.25\columnwidth}\centering\curve{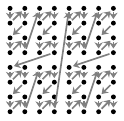}\vspace{-1.5mm}\caption{[1,0,1,0,1,0]}\vspace{1.5mm}\end{subfigure}\hfill
\begin{subfigure}{0.25\columnwidth}\centering\curve{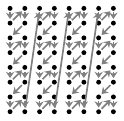}\vspace{-1.5mm}\caption{[1,0,1,1,0,0]}\vspace{1.5mm}\end{subfigure}\\
\begin{subfigure}{0.25\columnwidth}\centering\curve{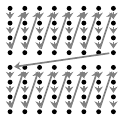}\vspace{-1.5mm}\caption{[1,1,0,0,0,1]}\vspace{1.5mm}\end{subfigure}\hfill
\begin{subfigure}{0.25\columnwidth}\centering\curve{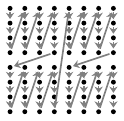}\vspace{-1.5mm}\caption{[1,1,0,0,1,0]}\vspace{1.5mm}\end{subfigure}\hfill
\begin{subfigure}{0.25\columnwidth}\centering\curve{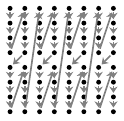}\vspace{-1.5mm}\caption{[1,1,0,1,0,0]}\vspace{1.5mm}\end{subfigure}\hfill
\begin{subfigure}{0.25\columnwidth}\centering\curve{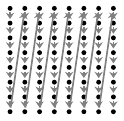}\vspace{-1.5mm}\caption{[1,1,1,0,0,0]}\vspace{1.5mm}\label{fig:layouts8x8:111000}\end{subfigure}
\caption{All 20 layouts for $8\times 8$ arrays generated by the family of indexing schemes described in \cref{sec:bijections}. Note that \cref{fig:layouts8x8:000111} corresponds to a row-major layout, while \cref{fig:layouts8x8:111000} corresponds to a column-major layout.}
\label{fig:layouts8x8}
\end{figure}

The Morton layout functions by interleaving the bits of the input indices in a fixed pattern: bits are drawn from each of the inputs in a round-robin manner. In this section, we generalize this idea, allowing bits to be interleaved in arbitrary order. This gives rise to more specialized layouts with different structure and, as a result, different extra-functional properties~\cite{doi:10.1177/1094342017725568,doi:10.1080/17445760902758560,10.1145/1274971.1274989}. \cref{fig:layouts8x8} shows all 20 layouts that are given by different bit interleaving orders for an $8 \times 8$ array. As with the standard Morton layout, the generalized Morton layout can be applied to any number of dimensions. As an example, the following three-dimensional layout selects two bits from the second index, one bit from the third index, then two bits from the first index, etc.:

\begin{equation}
\arraycolsep=1pt
\label{ex:permute:4}
f({\color{Palette1}011}_2, {\color{Palette2}101}_2, {\color{Palette3}100}_2) = \begin{array}{r r}&{\color{gray}0}{\color{Palette1}0}{\color{gray}00}{\color{Palette1}1}{\color{Palette1}1}{\color{gray}000}_2\\\vee&{\color{gray}000}{\color{Palette2}1}{\color{gray}000}{\color{Palette2}01}_2\\&{\color{Palette3}1}{\color{gray}0}{\color{Palette3}0}{\color{gray}000}{\color{Palette3}0}{\color{gray}00}_2\\\hline&{\color{Palette3}1}{\color{Palette1}0}{\color{Palette3}0}{\color{Palette2}1}{\color{Palette1}11}{\color{Palette3}0}{\color{Palette2}01}_2\end{array} = 313_{10}
\end{equation}

Our goal is to find Morton-like layouts i.e., bit-interleaving patterns, that improve application performance through an increase in cache efficacy. In this section, we will show that the design space for such layouts is very large, motivating the use of genetic algorithms. This necessitates a chromosomal representation of layouts, which we also present in this section. In addition, we describe how the canonical layouts can be described using the same representation, and we delve into practical considerations such as the computational cost of computing indices and support for same-instruction multiple-data (SIMD) processing.

\subsection{Enumerating Layouts}

\label{sec:bijections:enumerating}

We can characterize Morton-like layouts by the bit scattering pattern applied to each of the inputs (e.g., for \cref{ex:permute:4}, the first index is scattered to the fourth, fifth, and eighth bits). However, such a characterization is unsound in the sense that is allows us to describe invalid layouts: if two bits from any of the input indices are mapped onto the same bit in the output, the bitwise disjunction becomes an information-destroying operation and the layout becomes non-injective---that is, it would cause multiple multi-dimensional indices to map onto the same location in memory, making the layout unusable.

We can instead characterize layouts in a manner that is both complete and sound by enumerating the \emph{source} of each bit in the output index. In the remainder of this work we shall denote array layouts using sequences of the form $[i_0, \ldots, i_{n-1}]$, indicating the source indices in order of increasing bit significance: the least significant bit in the output index is drawn from the $i_0$th input index, the second-least significant bit is drawn from the $i_1$th input, and the most significant bit is drawn from the $i_{n-1}$th input. Note that each input bit must be used once and only once: whenever a bit is to be drawn from a given input index, we implicitly use the least significant bit for that input which has not yet been consumed. For the layout shown in \cref{ex:permute:4}, the two least significant bits are drawn from the second input, the third-least significant bit is drawn from the third input, and so forth: the resulting array layout is denoted using the sequence $[1,1,2,0,0,1,2,0,2]$. 

The aforementioned characterization of multi-dimensional layouts gives rise to families of layouts. The family of layouts over $n$ inputs, where each input has $b_0,\ldots,b_{n-1}$ bits, is isomorphic to the set of permutations of the multiset $S = \{0:b_0,\ldots,n-1:b_{n-1}\}$. We denote this set of permutations as $\mathfrak{S}(S)$. For convenience, we obviate the intermediate multiset such that $\mathfrak{S}'(b_0,\ldots,b_{n-1}) = \mathfrak{S}(\{0:b_0,\ldots,n-1:b_{n-1}\})$. We can then determine the total number of possible layouts as the number of multiset permutations of $\mathfrak{S}'$~\cite[p.~42]{brualdi1977introductory}:

\begin{equation}
\label{eq:permutation_set_size}
|\mathfrak{S}'(b_0, \ldots, b_{n-1})| = \binom{\sum_{i = 0}^{n-1} b_i}{b_0,\ldots,b_{n-1}} = \frac{\big(\sum_{i = 0}^{n-1} b_i \big)!}{\prod_{i = 0}^{n-1} (b_i!)}    
\end{equation}

\subsection{Including Canonical Layouts}

\label{sec:bijections:canonical}

It is worth noting that canonical layouts over arrays for which the size in each dimension is a power of two are, in fact, members of the family of Morton-like layouts. In order to sketch an informal argument for this, we recall that the indexing function for an $n$-dimensional canonical layout given array sizes $N_0, \ldots, N_{n-1}$ is defined as in \cref{eq:lexicographic}. If we assume that all sizes are powers of two, then the product of these sizes is guaranteed to be itself a power of two. Because multiplication by powers of two can be interpreted as a left-ward shift, the canonical layouts shift each input index $x_0, \ldots, x_n$ to a specific location in the binary expansion of the output index. Furthermore, because we assume $\forall i : x_i < N_i$, each bit in the output is determined by exactly one of the input indices; this allows us to interpret the summation as a series of bit-wise disjunctions, exactly like the definition of our Morton-like layouts. In general, a mode-0-major canonical layout of a $2^{b_0} \times \ldots \times 2^{b_{n-1}}$ array can be characterized---in the the scheme defined in \cref{sec:bijections:enumerating}---by contiguous subsequences of bits, each drawn from the same index i.e., a sequence of the following form:

\begin{equation}
\label{eq:cacnonical_repr}
[\underbrace{0,\ldots,0}_{b_0\text{ times}}, \underbrace{1,\ldots,1}_{b_1\text{ times}}, \ldots, \underbrace{n-1,\ldots,n-1}_{b_{n-1}\text{ times}}]    
\end{equation}

Canonical layouts with different major axes can be constructed by changing the order of the contiguous subsequences. The fact that the canonical layouts are members of the Morton-like family of array layouts allows us to evaluate the performance of these layouts in the exact same framework as the rest of the Morton-like layouts, and we will exploit this in \cref{sec:experiments}. 

\subsection{Hardware-Accelerated Indexing}

\label{sec:bijections:accel}

It is tempting to extend the aforementioned ideas to even more exotic indexing functions, like the Hilbert array layout~\cite{hilbert1891ueber,10.1145/3555353,10.1007/978-3-642-25100-9_73}. The computational cost of many such functions renders them impractical, however: if the cost of computing memory addresses is too large, any performance gained by improving the cache-friendliness of a program will be negated. The Morton-like layouts we consider in this work allow efficient index calculations on modern commodity hardware, which we demonstrate in this section.

Under canonical array layouts, indices are calculated either iteratively through repeated addition and multiplication, or in parallel through parallel multiplication followed by reduction through addition. In $n$-dimensional cases both approaches require $n-1$ additions and $n-1$ multiplications, operations which can be efficiently performed on virtually all processors. Specifically, the Intel Haswell and AMD Zen 3 microarchitectures---on which we focus in this work---can perform 64-bit register addition (\texttt{ADD r64 r64}) with a latency 1 cycle and a reciprocal throughput of 0.25 cycles, while they can execute multiplication (\texttt{IMUL r64 r64}) with a latency of 3 cycles and a reciprocal throughput of 1 cycle~\cite{10.1145/3297858.3304062}.

Our bit-interleaving array layouts rely, in $n$-dimensional cases, on $n-1$ bitwise disjunctions and $n$ bit-scatter operations. Such disjunctions (\texttt{OR r64 r64}) can be performed with a latency of 1 cycle and a reciprocal throughput of 0.25 cycles---the same as the \texttt{ADD} instruction---on both of the aforementioned microarchitectures. We perform the bit-scattering operation using the \emph{parallel bit deposition} (\texttt{PDEP r64 r64 r64}) instruction, which is included in the BMI2 extension to the x86-64 instruction set~\cite{10.1145/3319647.3325833}. The Intel Haswell and AMD Zen 3 microarchitectures both perform bit deposition with a latency of 3 cycles and a reciprocal throughput of 1 cycle, identical to the \texttt{IMUL} instruction. It follows that Morton-like indexing requires---in theory---only a single additional instruction over canonical index calculation.

The hardware extension required to perform bit deposition is widely supported: BMI2 has been included in Intel processors starting with the Haswell microarchitecture (2013)~\cite{6762795}, and in AMD processors starting with the Excavator microarchitecture (2015), albeit in a limited fashion; AMD processors gained full hardware support for these instructions starting with the Zen 3 microarchitecture (2020)~\cite{9718180}\footnote{Pre-Zen 3 processors supported parts of the BMI2 instruction set---the \texttt{PEXT} and \texttt{PDEP} instructions in particular---through emulation in microcode rather than in hardware, making them very slow.}. 

\begin{figure}
    \centering
    \includegraphics{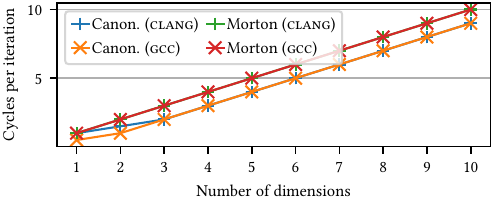}
    \caption{Throughput of a kernel calculating array indices using canonical layouts as well as Morton-like layouts on the Intel Haswell microarchitecture as given by \textsc{OSACA}.}
    \label{fig:canon_vs_morton}
\end{figure}

In order to further evaluate the competitiveness of Morton-like layouts compared to canonical layouts, we analyze implementations of both indexing schemes over a range of dimensionalities as compiled by \textsc{gcc} 12.3 and \textsc{clang} 15.0 using \textsc{OSACA} 0.5.2~\cite{8641578}. All code was compiled using the \texttt{-O2} optimization flag. The results of this analysis are shown in \cref{fig:canon_vs_morton}. Over the range of dimensionalities considered, the canonical layouts are consistently faster i.e., require fewer cycles to compute, than the Morton-like layouts. However, the difference in performance---approximately one cycle---is relatively small and overshadowed by the number of cycled saved due to a reduction in cache misses. Furthermore, we focus primarily on memory-bound applications, in which a small increase in index calculation time is unlikely to affect performance. We conclude, therefore, that Morton-like layouts are competitive with canonical layouts strictly in terms of address computation costs.

\subsection{Support for SIMD}

\label{sec:bijections:SIMD}

An important consideration in the design of array layouts is the ability to vectorize kernels through single-instruction multiple-data (SIMD) operations. Canonical layouts guarantee the contiguity of fibers in the array, which facilitates the (automated) vectorization (e.g., the application of SIMD) of many operations, and this benefit is lost when applying the array layouts discussed in this paper. However, we posit that there remains ample opportunity to accelerate computation on Morton-like arrays using SIMD, and we argue this by distinguishing two classes of computation patterns.

The first class consists of \emph{unstructured} patterns in which data is operated on element-wise without spatial context i.e., without consideration of nearby elements; a prominent example of such an operation is matrix \emph{addition}. In such applications, SIMD can be trivially applied to the underlying one-dimensional memory, regardless of the layout of the data: since elements can be added point-wise in any order, doing so in the order in which the data is laid out in memory is both feasible and enables SIMD.

The second class of problems consists of \emph{structured} patterns in which operations must be performed in a specific order. A prime example of such an operation is matrix \emph{multiplication} where the inner product of fibers must be computed. In such cases, it is imperative that fibers can be accessed in contiguous blocks. The size of these blocks depends on the vectorization technology used as well as the size of the data type: in the x86 instruction set, SSE vectorisation requires two consecutive double-precision numbers or four consecutive single-precision numbers~\cite{intelx86}; the much wider ARM SVE instruction set extension~\cite{7924233} may require up to thirty-two consecutive double-precision numbers or sixty-four single-precision numbers.

In order to facilitate vectorization for structured patterns of computation, we can impose certain constraints on the array layouts we consider. Indeed, if the $n$ least-significant bits of an interleaving pattern are all drawn from the $m$th input index, then the layout guarantees that the mode-$m$ fibers in the array are contiguous in blocks of $2^n$ elements. This requirement can be incorporated into the selection of array layouts; for example, we can enable efficient AVX2 vectorisation (with a vector width of 256 bits) using single-precision (32-bit) floating point numbers by ensuring that the three least significant bits in an array layout are drawn from the same source. In other words, we can easily constrain our search space to include only array layouts with properties that favor vectorization, and we believe that doing so will enable SIMD-accelerated computation arrays laid out in Morton-like orders.

\section{Exploration Through Evolution}

\label{sec:exploration}

The canonical set of indexing bijections for laying out multi-di\-men\-sion\-al memory is small: for two-dimensional data, there are two possible layouts, and the performance of these layouts can be evaluated using exhaustive benchmarks~\cite{10.1145/3578244.3583723,10.1002/cpe.1018,814600}. Exhaustively exploring the family of indexing function outlined in \cref{sec:bijections}, however, is impractical owing to the sheer number of permissible permutations. Importantly, the number of canonical layouts increases only with the number of \emph{dimensions}, while the number of Morton-like layouts increases with both the number of dimensions and the \emph{size} of the array in each of those dimensions. By \cref{eq:permutation_set_size}, a small $4 \times 4$ array (indexed by two bits in each dimension) can be laid out in $\nicefrac{(2+2)!}{2!2!} = 6$ ways. A larger array of size $\num{4096} \times \num{4096}$ (twelve bits in each dimension) can be laid out in $\nicefrac{(12+12)!}{12!12!} = \num{2704156}$ ways. A three-dimensional array of size $256 \times 256 \times 256$ has the same number of elements as the aforementioned $\num{4096} \times \num{4096}$ array, but permits $\nicefrac{(8+8+8)!}{8!8!8!} = \num{9465511770}$ permutations. As these examples indicate, the number of possible permutations quickly scales beyond what can be feasibly explored through exhaustive search; in order to tackle the explosive growth in the design space for Morton-like layouts, we propose the use of genetic algorithms (\cref{sec:backgroun:genetic}).

\subsection{Genetic Algorithm Configuration}
In this work, we employ a relatively simple $(\lambda, \mu)$-ES genetic algorithm~\cite{holland1992adaptation,Slowik2020}. The chromosomal representations of array layouts is identical to the characterization given in \cref{sec:bijections:enumerating}, and this gives rise to a combinatorial optimization problem. We facilitate the recombination of array layouts into novel layouts using the ordered crossover (OX) operator~\cite{10.5555/1625135.1625164}, and we employ inversion-based mutation~\cite{Eiben2015}. Our approach differs from classical genetic algorithms in only one significant way: our initial population is not chosen randomly from the solution space. Instead, the initial populations for our evolutionary experiments always consist of two individuals, depicting two canonical layouts for a given array size, as described in \cref{sec:bijections:canonical}. We choose to do this to ensure that our initial populations are unbiased and deterministic, allowing us to more easily assess the efficacy of our genetic strategy.

\subsection{Fitness Function Design}

\label{sec:exploration:fitness}

There are two general strategies for evaluating the performance i.e., fitness, of a given array layout under a given cache hierarchy and access pattern: measurement and simulation. In order to assess fitness through \textit{measurement}, we execute a program on actual hardware and measuring the running time of the process. Although such a fitness function is conceptually simple, it suffers from two primary flaws:
\begin{enumerate*}
    \item measurements are noisy and may suffer from run-to-run variance, which may hinder the performance of genetic algorithms~\cite{miller1995genetic}---in particular, our genetic algorithm is vulnerable to noise stemming from cache pollution effects; and
    \item measurements require access to the target hardware, which may be inconvenient or even impossible---for example, in hardware-software co-design scenarios, where the hardware under consideration does not (yet) exist.
\end{enumerate*}
For these reasons, we choose not to base our fitness function on measurements. 

Instead, we employ \emph{simulation} for which we need a simulator that can accurately compare the cache performance for different access-patterns on the same cache hierarchy. 
For this, we selected \textsc{pycachesim}, a component of the \textsc{Kerncraft} toolkit~\cite{10.1007/978-3-319-56702-0_1}
. We use \textsc{pycachesim} by simulating an access pattern such as matrix multiplication and registering the relevant trace of load and store operations. After all accesses have been recorded, we force a write-back of the caches and collect the number of hits and misses in each cache level. We combine the number of hits in every cache level as well as in main memory with the latency of retrieving data from each of these levels to compute the total number of cycles spent retrieving data from the cache hierarchy. Given an array layout $I$, an access pattern $A$ and a simulated cache hierarchy $H$, we calculate the total number of cycles using the following equation, in which $\mathrm{L}i_\mathrm{hit}$, $\mathrm{L}i_\mathrm{miss}$, and $\mathrm{L}i_\mathrm{lat}$ represent the number of hits, the number of misses, and the latency of the $i$th cache level%
, and $M$ represents main memory:

\begin{equation}
\label{eq:fitness-pre}
    C(I; A, H) = \mathrm{M}_\mathrm{hit}(I; A, H)\mathrm{M}_\mathrm{lat}(H) + \sum_i \mathrm{L}i_\mathrm{hit}(I; A, H)\mathrm{L}i_\mathrm{lat}(H)
\end{equation}

From this, we compute an approximation of the number of accesses performed per cycle, giving rise to a higher-is-better fitness function defined as follows:

\begin{equation}
\label{eq:fitness}
F(I; A, H) = \frac{\mathrm{L1}_\mathrm{hit}(I; A, H) + \mathrm{L1}_\mathrm{miss}(I; A, H)}{\mathrm{L1}_\mathrm{lat}(H)\cdot C(I; A, H)}
\end{equation}

Intuitively, the numerator in \cref{eq:fitness} counts the total number of memory accesses, as all accesses either hit or miss in L1. 
The denominator, then, estimates the total number of cycles spent retrieving data from the various cache levels. The denominator is multiplied by a normalizing factor equal to the latency of the L1 cache; it follows from \cref{eq:fitness-pre} that the achievable performance is softly bound by the reciprocal of the L1 access latency. Indeed, this performance is achieved if and only if all accesses hit the L1 cache. Normalizing the fitness function using the L1 cache latency improves our ability to compare results between different cache hierarchies.

\section{Evaluation}

\label{sec:experiments}

We evaluate the efficacy of the methods hitherto discussed by demonstrating that  
\begin{enumerate*}
    \item our fitness function is well-chosen i.e., that is correlates with performance measurements in real hardware; that
    \item our evolutionary process is capable of finding novel array layouts with favorable cache properties; and that
    \item the layouts which are found by our evolutionary process actually lead to relevant performance gains in real hardware.
\end{enumerate*}
Our validation is based on eight distinct access patterns and two processors with distinct cache hierarchies.

\subsection{Experimental Setup}

\begin{lstfloat}
\begin{lstlisting}[style=mystyle,frame=tlrb,basicstyle={\scriptsize\ttfamily},language=c++]
template <concepts::array<2> M>
void mm_ijk(const M & A, const M & B, M & C) {
    const auto m = C.get_size();
    for (std::size_t i = 0; i < m; ++i) {
        for (std::size_t j = 0; j < m; ++j) {
            typename M::value_type acc = 0.;
            for (std::size_t k = 0; k < m; ++k)
                acc += A.load(i, k) * B.load(k, j);
            C.store(acc, i, j);
        }
    }
}
\end{lstlisting}\vspace{-2mm}
\caption{Example of how an access pattern (\textsc{MMijk}) is described in C++. Metaprogramming allows the same source to be used for both simulation and execution on real hardware.}
\label{lst:mmijk_example}
\end{lstfloat}

We consider a set of eight access patterns loosely based on the selection of algorithms used by \citeauthor{10.1002/cpe.1018}~\cite{10.1002/cpe.1018}. The access patterns were picked to represent common real-world applications (dense linear algebra and fluid dynamics), to represent both two-dimensional and three-dimensional applications, and to differ in critical properties such as memory size and number of loads and stores. A description of the access patterns we consider in this paper is given in \cref{tab:pattern_stats}.

All our access patterns are described using C++ code---see the example in \cref{lst:mmijk_example}---which ensures high performance as opposed to the Python code used for our evolutionary processes; the interaction between the C++ and Python components of our project is managed using \textsc{pybind11}~\cite{pybind11}. We use template meta-programming to generalize our access patterns in such a way that a single definition can be used for both simulation and benchmarking without loss of performance due to run-time polymorphism; this eliminates any possible discrepancies between the code used for simulation and the code used for measurement.

\begin{table*}
    \centering
    \caption{Overview of the access patterns used for evaluation, including the use of memory and the number of loads and stores.}
    {\small
    \begin{tabular}{l p{8.7cm} r r r}
        \toprule
        Access pattern & Description & Mem.\ size & Loads & Stores  \\\midrule
        $\textsc{MMijk}(m; s)$ & Multiplication of two $2^m \times 2^m$ matrices, both of $s$-byte real numbers. & $3\cdot s \cdot 2^{2m}\,\si{\byte}$ & $2 \cdot 2^{3m}$ & $2^{2m}$ \\
        $\textsc{MMTijk}(m, n; s)$ & Multiplication of a $2^m \times 2^n$ matrix by a transposed $2^m \times 2^n$ matrix. & $s \cdot (2 \cdot 2^{m+n} + 2^{2n}) \si{\byte}$ & $2\cdot2^{2m+n}$ & $2^{2m}$ \\
        $\textsc{MMikj}(m; s)$ & Same as $\textsc{MMijk}(m;s)$ with the order of the inner loops switched. & $3\cdot s \cdot 2^{2m}\,\si{\byte}$ & $3 \cdot 2^{3m}$ & $2^{3m}$ \\
        $\textsc{MMTikj}(m, n; s)$ & Same as $\textsc{MMTijk}(m, n; s)$ with the order of the inner loops switched. & $s \cdot (2 \cdot 2^{m+n} + 2^{2n}) \si{\byte}$ & $3\cdot2^{2m+n}$ & $2^{2m+n}$ \\
        $\textsc{Jacobi2D}(m, n; s)$ & Four-point stencil kernel over a $2^m \times 2^n$ array of $s$-byte real numbers. & $2 \cdot s\cdot 2^{m+n}\,\si{\byte}$ & $\sim4 \cdot 2^{m + n}$ & $2^{m + n}$\\
        $\textsc{Cholesky}(m; s)$ & Cholesky--Banachiewicz decomposition of a $2^m \times 2^m$ matrix. & $2 \cdot s \cdot 2^{2m}\,\si{\byte}$ & $2 \cdot 2^{2m}$ & $\sim\frac{1}{2} \cdot 2^{2m}$\\
        $\textsc{Crout}(m; s)$ & Crout decomposition of a $2^m \times 2^m$ matrix of $s$-byte real numbers. & $2 \cdot s \cdot 2^{2m}\,\si{\byte}$ & $\frac{7}{2} \cdot 2^{2m}$ & $2^{2m}$ \\
        $\textsc{Himeno}(m, n, p; s)$ & Nineteen-point Himeno stencil \cite{himeno} over $2^m \times 2^n \times 2^p$ arrays. & $12\cdot s \cdot 2^{m + n + p}\,\si{\byte}$ & $24 \cdot 2^{m+n+p}$ & $2^{m+n+p}$\\
         \bottomrule
    \end{tabular}}
    \label{tab:pattern_stats}
\end{table*}


\label{sec:results:simulated_hardware}

\begin{lstfloat}
\begin{sublstfloat}[b]{0.49\columnwidth}
\begin{lstlisting}[style=mystyle,frame=tlrb,basicstyle={\scriptsize\ttfamily}]
caches:
  L1:
    sets: 64
    ways: 8
    line: 64
    replacement: LRU
    write_back: true
    store_to: L2
    load_from: L2
    latency: 4
  L2:
    sets: 512
    ways: 8
    line: 64
    replacement: LRU
    write_back: true
    store_to: L3
    load_from: L3
    victim_to: L3
    latency: 12
  L3:
    sets: 25600
    ways: 16
    line: 64
    replacement: LRU
    write_back: true
    latency: 36
memory:
  first: L1
  last: L3
  latency: 200
\end{lstlisting}\vspace{-2mm}
\caption{Intel Xeon E5-2660 v3}
\label{lst:cache_spec:haswell}
\end{sublstfloat}
\begin{sublstfloat}[b]{0.49\columnwidth}
\begin{lstlisting}[style=mystyle,frame=tlrb,basicstyle={\scriptsize\ttfamily}]
caches:
  L1:
    sets: 64
    ways: 8
    line: 64
    replacement: LRU
    write_back: true
    store_to: L2
    load_from: L2
    latency: 7
  L2:
    sets: 1024
    ways: 8
    line: 64
    replacement: LRU
    write_back: true
    store_to: L3
    load_from: L3
    victim_to: L3
    latency: 12
  L3:
    sets: 32768
    ways: 16
    line: 64
    replacement: LRU
    write_back: true
    latency: 46
memory:
  first: L1
  last: L3
  latency: 200
\end{lstlisting}\vspace{-2mm}
\caption{AMD EPYC 7413}
\label{lst:cache_spec:zen3}
\end{sublstfloat}
\caption{Two examples of cache specifications for different CPU models. Note that these configurations are approximations of the true cache hierarchies.}
\label{lst:cache_spec}
\end{lstfloat}

We conduct our experiments on two different CPUs: the Intel Xeon E5-2660~v3~\cite{e52660} based on the Haswell microarchitecture~\cite{6762795}, and the AMD~EPYC~7413~\cite{epyc7413} based on the Zen~3 microarchitecture~\cite{9718180}. When we perform experiments on non-simulated Haswell processors we use the \blind{the DAS-6 cluster~\cite{7469992}}, whereas we use \blind{a machine located at CERN} for experiments on Zen~3 processors. When we perform experiments based on simulation, we use the \blind{the DAS-6 cluster~\cite{7469992}} and configure our cache simulator according the cache configurations shown in \cref{lst:cache_spec:haswell} for the Haswell processor, and \cref{lst:cache_spec:zen3} for the Zen~3 processor. Note that the cache configurations are based on the accessibility of caches from a single core. This is especially relevant for the L3 cache on the Zen~3 chip, which is shared across groups of cores rather than the entire CPU: in the case of the AMD~EPYC~7413, the CPU comes equipped with \SI{128}{\mebi\byte} of L3 cache, but only \SI{32}{\mebi\byte} is accessible from any single core~\cite{9718180}. We simplify the cache replacement policies of the actual hardware by assuming LRU caches (i.e., caches with a least-recently-used eviction policy); in reality, the Haswell caches employ eviction policies consistent with tree-PLRU (tree-based pseudo-LRU) for the L1 and L2 caches~\cite{10.1145/3385412.3386008,10.1145/3297858.3304062}, while the L3 cache is consistent with a set-dueling-controlled adaptive insertion policy~\cite{4460516,10.1145/3297858.3304062}. Cache sizes were gathered from specification documents~\cite{amd19h,6762795}, while cache latencies were obtained optimistically from sources on the fastest load-to-use latencies~\cite{amd19h,hwbp}. The Zen~3 L1 cache has a fastest load-to-use latency of four cycles for integers and seven cycles for floating point values~\cite{amd19h}---we use the latter in our simulations. Finally, we assume a constant 200 cycle access latency for main memory in both systems.

\subsection{Fitness Function Validation}

\label{sec:res:fitval}

The fitness function we use in our evolutionary process (\cref{sec:exploration:fitness}) is based on simulation results because 
simulation yields significant benefits over empirical measurements, primarily in terms of determinism and in the ability to simulate future hardware. However, this strategy is not without risk: the simulation we perform is based on a non-cycle-accurate simulator, uses simplified cache hierarchies, and ignores computation entirely. 
Consequently, we must evaluate the usefulness of our fitness function by establishing its correlation with execution time in real hardware.

Ideally, the running time of a kernel using a given array layout would correlate inversely linearly with our fitness function, therefore ensuring two important properties. 
Firstly and most importantly, it guarantees that running time decreases monotonically with the value of the fitness function, such that an array layout with a higher fitness value is guaranteed to run more quickly; this allows us to establish a ranking of layouts and enables us to reliably select the best-performing array layout. 
Secondly, linear correlation guarantees proportionality between fitness and running time, which facilitates the weighted selection of individuals.

To evaluate the degree to which the aforementioned criteria are met, we randomly select one hundred array layouts for each of the eight access patterns given in \cref{tab:pattern_stats}. We then evaluate the simulated fitness and measure the running time in real hardware of each pair of array layout and access pattern. The fitness functions of the pairs are calculated in parallel, as they are designed to be deterministic and impervious to cache pollution or resource contention. The empirical benchmarks are performed sequentially, ensuring that the benchmark is the sole user of the processor caches. All measurements are repeated ten times, and we report the mean and standard deviation of the running time.

\begin{figure}
    \centering
    \includegraphics{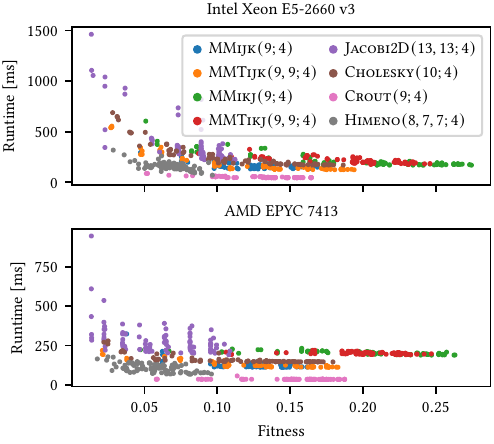}
    \caption{Scatter plot of the fitness and measured running time on an Intel Xeon E5-2660 v3 CPU and AMD EPYC 7413 for randomly chosen array layouts.}
    \label{fig:fitness_corr}
\end{figure}

The results of this experiment are shown in \cref{fig:fitness_corr}. The coefficient of variation of the measurements never exceeded a value of $c_{\mathrm{v}} = 0.0801$. Accordingly, we have opted to omit error bars from the figure. Upon visual inspection, it is clear that the correlation between our fitness function and running time is not linear, although the two do appear correlated. We confirm our suspicions of correlation by computing Pearson's coefficient of correlation ($\rho_p$) and Spearman's coefficient of rank correlation ($\rho_s$); the resulting statistics are given in \cref{tab:correlation_table}. We observe that our fitness function and running time correlate moderately to strongly with running time for the Intel Xeon E5-2660 v3 processor, although the correlation is weaker for the AMD EPYC 7413 processor. Although it is clear that there is space for the fitness function to be improved, we believe that it correlates sufficiently with running time to enable its use in genetic algorithms.

\begin{table}
    \centering
    \caption{Pearson's coefficient of correlation ($\rho_p$) and Spearman's coefficient of rank correlation ($\rho_s$) between our simulation-based fitness function and true running time.}
    \begin{tabular}{l R{1cm} R{1cm} R{1cm} R{1cm}}
        \toprule
        & \multicolumn{2}{c}{Intel E5-2660 v3} & \multicolumn{2}{c}{AMD EPYC 7413} \\\cmidrule(lr){2-3}\cmidrule(lr){4-5}
        Access pattern & $\rho_{p}$ & $\rho_{s}$ & $\rho_{p}$ & $\rho_{s}$ \\\midrule
        $\textsc{MMijk}(9; 4)$ & \num{-0.672} & \num{-0.480} & \num{-0.648} & \num{-0.489} \\
        $\textsc{MMTijk}(9, 9; 4)$ & \num{-0.810} & \num{-0.896} & \num{-0.863} & \num{-0.823} \\
        $\textsc{MMikj}(9; 4)$ & \num{-0.845} & \num{-0.815} & \num{-0.800} & \num{-0.838} \\
        $\textsc{MMTikj}(9, 9; 4)$ & \num{-0.777} & \num{-0.744} & \num{-0.291} & \num{-0.405} \\
        $\textsc{Jacobi2D}(13, 13; 4)$\hspace{-2mm} & \num{-0.760} & \num{-0.769} & \num{-0.390} & \num{-0.428} \\
        $\textsc{Cholesky}(10; 4)$ & \num{-0.827} & \num{-0.953} & \num{-0.725} & \num{-0.892} \\
        $\textsc{Crout}(9; 4)$ & \num{-0.846} & \num{-0.663} & \num{-0.213} & \num{-0.704} \\
        $\textsc{Himeno}(8, 7, 7; 4)$ & \num{-0.607} & \num{-0.475} & \num{-0.561} & \num{-0.496} \\
        \bottomrule
    \end{tabular}
    \label{tab:correlation_table}
\end{table}

\subsection{Genetic Algorithm Performance}

\label{sec:expr:evolution}
To evaluate our evolutionary process (\cref{sec:exploration}) as a whole,  
we intend to verify that it can, indeed, find Morton-like array layouts that have a higher simulated fitness than the canonical layouts. To this end, we perform the evolutionary process for each combination of our two simulated processors and eight access patterns, giving rise to a total of sixteen experiments. For all of these experiments, we configure our genetic algorithm to use $\mu = 20$, $\lambda = 20$, and a mutation rate of $25\%$. We simulate a total of \num{20} generations in each case.

\begin{figure}
    \centering
    \includegraphics{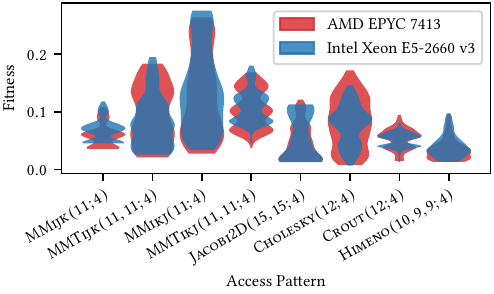}
    \caption{Distribution of the fitness values for all individuals found across evolution experiments for eight access patterns and two processors.}
    \label{fig:fitness_violin}
\end{figure}

\begin{figure}
    \centering
    \includegraphics{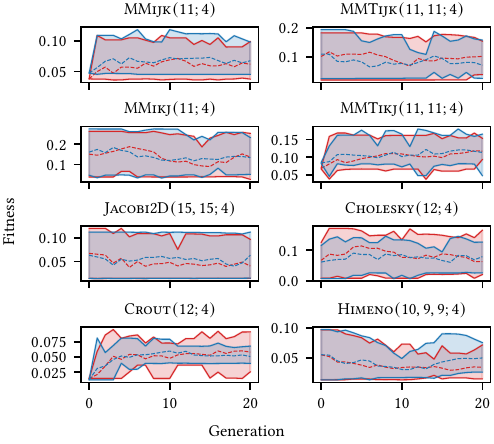}
    \caption{Range of fitness values across eight experiments for the Intel Xeon E5-2660 v3 (blue) and AMD EPYC 7413 (red). Mean fitness values are given by the dashed lines.}
    \label{fig:fitness_evolution}
\end{figure}

\cref{fig:fitness_violin} shows a violin plot of the fitness distribution of all individuals considered during the evolutionary process. \cref{fig:fitness_evolution} shows the evolution of population fitness over the course of our experiments. Note that each of these experiments represents a single evolutionary process. We notice that for the \textsc{MMTijk}, \textsc{MMikj}, \textsc{Jacobi2D}, and \textsc{Himeno} access patterns, our method does not manage to discover any layouts with higher fitness than the initial population of canonical layouts. In the experiment on the \textsc{MMijk} access pattern, we discover layouts with a fitness $149.8\%$ higher than the canonical layouts on the Intel Xeon E5-2660 v3 processor, and we improve on the fitness of canonical layouts by $187.5\%$ for the AMD EPYC 7413. We also find layouts with improved fitness for the \textsc{MMTikj} ($109.6\%$ and $141.1\%$ for the Intel and AMD processors, respectively), \textsc{Cholesky} ($26.4\%$ and $36.8\%$), and \textsc{Crout} ($545.9\%$ and $541.1\%$) access patterns. It is notable that we are able to find layouts with high fitness in few generations.

\subsection{Real-World Performance}

\label{sec:results:real-world}

In order to evaluate whether the layouts identified by our evolutionary algorithms as superior to canonical layouts are indeed better, we evaluate them on real hardware. We collect the fittest individual from each of the successful evolution experiments---i.e., experiments in which our method improved upon canonical layouts, as indicated by the top boundary in \cref{fig:fitness_evolution} exceeding the maximum fitness in the first generation---and evaluate the performance of those layouts compared to the canonical layouts on real hardware. Given that our genetic algorithm discovered superior layouts for four access patters---\textsc{MMijk}, \textsc{MMTikj}, \textsc{Cholesky}, and \textsc{Crout}--and that we evaluate a discovered layout and two canonical layouts for each access pattern, this gives rise to twenty-four experiments. We repeat each experiment ten times to compensate for run-to-run variance.

The results of our experiments are shown in \cref{tab:speedup_table}; they show that some access patterns---the \textsc{Cholesky} pattern in particular---benefit very little from our method, with speed-ups ranging from small on the Haswell processor to insignificant on the Zen 3 processor. The matrix multiplication access patterns benefit more, and performance for these access patterns is improved significantly. The \textsc{Crout} access pattern stands out as achieving very large speedup---up to a factor ten---from our method. It is worth noting that, in most cases, the Zen 3 processor benefits more from our evolutionary methodology than the Haswell processor; we do not currently have a satisfactory explanation for this behavior.

\begin{table}
    \centering
    \caption{Comparison of running time between the best-performing canonical layout and the best-performing layout found by our evolutionary process for four access patterns.}
    \begin{tabular}{l l S[table-format=3.2,table-space-text-post = \,\si{\second}] S[table-format=3.2,table-space-text-post = \,\si{\second}] r}
    \toprule
    \multicolumn{2}{l}{Access pattern} & {Best can.} & {Best evo.} & {Speedup} \\\midrule
    \multicolumn{2}{l}{Intel Xeon E5-2660 v3} & \\
    & $\textsc{MMijk}(11; 4)$ & 17.84\,\si{\second} & 10.94\,\si{\second} & $63.1\%$ \\
    & $\textsc{MMTikj}(11, 11; 4)$ & 18.13\,\si{\second} & 13.96\,\si{\second} & $29.9\%$ \\
    & $\textsc{Cholesky}(12; 4)$ & 11.84\,\si{\second} & 11.43\,\si{\second} & $3.6\%$ \\
    & $\textsc{Crout}(12; 4)$ & 158.54\,\si{\second} & 43.72\,\si{\second} & $262.6\%$ \\
    \multicolumn{2}{l}{AMD EPYC 7413} & \\
    & $\textsc{MMijk}(11; 4)$ & 37.71\,\si{\second} & 9.58\,\si{\second} & $293.8\%$ \\
    & $\textsc{MMTikj}(11, 11; 4)$ & 32.35\,\si{\second} & 15.21\,\si{\second} & $112.6\%$ \\
    & $\textsc{Cholesky}(12; 4)$ & 9.72\,\si{\second} & 9.55\,\si{\second} & $1.0\%$ \\
    & $\textsc{Crout}(12; 4)$ & 232.84\,\si{\second} & 21.03\,\si{\second} & $1007.0\%$ \\
    \bottomrule
    \end{tabular}
    \label{tab:speedup_table}
\end{table}

It is important to note that we do not claim to have discovered a novel way of performing matrix multiplication or matrix decomposition that outperforms existing implementations. Indeed, our experiments are based on relatively naive implementations of these algorithms; high-performance implementations of matrix multiplication commonly rely on tiling to significantly improve the cache behavior of the application~\cite{1214317}, and the performance of tiled matrix multiplication surpasses what we achieve in this paper. The purpose of the methodology described in this paper, rather, is to provide an alternative way of improving the cache behavior of an application in a manner which is fully agnostic of the application: unlike tiling and other application-specific optimizations, our methodology of altering the array layouts can be applied to any multi-dimensional problem without the need for application-specific knowledge. In addition, our approach requires few code changes, making it easy to implement.
\section{Limitations and Threats to Validity}

\label{sec:threats}

Throughout this work, we evaluate cache efficacy through a simplified lens which may reduce the applicability of our methods in more complex, real-world applications. Indeed, we consider accesses to memory in isolation, decoupled from computation and cache-polluting effects. We assume single-threaded execution without scheduling, which means that our caches will not be polluted by processes sharing (parts of) the cache hierarchy, nor will the application have its cached data evicted due to context switching. We also assume scalar, in-order execution of memory accesses. Finally, we take an optimistic view of cache latencies, using the fastest load-to-use latencies provided by hardware manufacturers; in real-world scenarios, cache latencies may be both more pessimistic and less stable than we assume. The results shown in \cref{sec:results:real-world} indicate, however, that our fitness function is sufficiently accurate to be effective in real hardware. 

In addition, the family of array layouts described in this work requires array sizes to be powers of two in each dimension. In applications where this is not the case, arrays must be over-allocated. For $n$-dimensional applications, using the layouts described in this paper requires over-allocation by a factor of $\mathcal{O}(2^n)$. Furthermore, applications using such layouts must consider the use of SIMD vectorization: it remains an open question which operations on arrays laid out in non-standard ways can be (automatically) vectorized. We have argued for the feasibility of SIMD in Morton-like arrays in \cref{sec:bijections:SIMD}.

Finally, our work considers only multiset permutations, in which the rank significance of bits in the input indices is preserved. This decision is based on current commodity hardware, which is capable of efficiently permuting bits only under this condition. There exists an even larger family of layouts in which rank bit significance is not preserved\footnote{That is to say, the layout $[0_0, 0_1, 1_0, 1_1]$ (which draws its least significant bit from the least significant bit of the first index) is distinct from the layout $[0_1, 0_0, 1_0, 1_1]$ (which instead draws its least significant bit from the \emph{second-least} significant bit of the first index).}; such layouts could be of practical use in theoretical future processors with more advanced bit manipulation instructions, or in current FPGA and ASIC devices which permit the implementation of custom bit manipulation operations. Although we have not tested our approach on this further generalization, we are confident that an evolutionary approach like the one presented in this paper could be beneficial in exploring this (even larger) design space.

\section{Reproducibility and Reusability}

The evolutionary algorithms, scripts for the processing and visualisation of data, and other software used in this paper are permanently archived on Zenodo~\cite{swatman_2024_10567243}, and have been made available at \doi{10.5281/zenodo.10567243}. The aforementioned artifact also contains all data that was gathered and processed during the work presented in this paper. For more information about the use of the artifact accompanying this paper, see the included \texttt{README} file.

\section{Conclusions and Future Work}

\label{sec:conclusion}

In this paper, we have discussed a generalization of the Morton layout for multi-dimensional data and we have shown that there exist families of array layouts with strongly varying cache behavior which, in turn, impact the performance of applications. We have shown how these layouts can be systematically described, and that the number of possible layouts quickly exceed the limits of what can be feasibly explored using exhaustive search. We have proposed a method based on evolutionary algorithms for the exploration of the design space of such layouts. We have evaluated the fitness of different array layouts using cache simulation and we have presented results indicating that our fitness function correlates with real world performance. Furthermore, we have shown that the methodology described in this paper can be used to improve the performance of applications on real hardware by up to ten times.

In the future, we intend to investigate the use of multi-objective optimization using NSGA-II~\cite{996017} in order to find array layouts that provide favorable cache behavior across multiple applications. We also intend to explore more advanced genetic algorithms which are known to perform well in combinatorial problems, such as RKGA~\cite{doi:10.1287/ijoc.6.2.154} and BRKGA~\cite{Goncalves2011}. It is our belief that exploring more evolutionary strategies will give us more insight into the convergence properties of various methods, and allow us to select the most efficient one. Although our fitness function correlates with real-world performance, the correlation is not perfect; we believe that the efficacy of our method could be improved through the development of more advanced fitness function, perhaps through the use of machine learning methods. In particular, we believe that the field of metric learning may enable us to develop more accurate fitness functions, and we aim to explore this avenue of research in the future. Finally, we aim to expand our research to a broader range of access patterns and hardware, including graphics processing units (GPUs).

\begin{acks}
The work presented in this paper was done in the context of the CERN Doctoral Student Programme. Many of the experimental results shown in this paper were gathered on the Advanced School for Computing and Imaging (ASCI) DAS-6 compute cluster~\cite{7469992}.
\end{acks}

\bibliographystyle{ACM-Reference-Format}
\bibliography{main}
\end{document}